# Vandalism Detection in Wikipedia: a Bag-of-Words Classifier Approach

Amit Belani
ab422@cornell.edu
November 11, 2009

## Abstract

A bag-of-words based probabilistic classifier is trained using regularized logistic regression to detect vandalism in the English Wikipedia. Isotonic regression is used to calibrate the class membership probabilities. Learning curve, reliability, ROC, and cost analysis are performed.

## 1. Motivation

Wikipedia is a collaborative encyclopedia that anyone can edit and improve. Although it has been argued that this openness is one of the reasons for its success, it does allow some unscrupulous editors to introduce damaging edits. At the present time, most of these edits are identified and reverted either by human editors or by automated anti-vandal bots.

Prominent anti-vandal bots such as ClueBot and VoABot II use lists of regular expressions and user blacklists to detect vandalism. These regular expression rules are created manually and are difficult to maintain. They detect only 30% of the committed vandalism. Their combined Recall was observed to be 33%, with a Precision of 100%.[1]

Since 2008, machine learning algorithms have been attempted for this task,[1-5] but with limited success, leaving room for improvement. Moreover, the methods attempted have either utilized only a small sample of the available training data, or have generally not considered a cost-sensitive approach to model development. Such an approach aids in improving the true positives detection rate while maintaining low false positives. It is believed that a false positive would have a higher cost than a false negative in any practical application of the learning algorithm for this classification problem.

## 2. Task

An edit is a sequential revision to an article. A *revert* is an edit that returns the article to a previous version. That is, for versions $i, j, k$, where $i < j < k$ in chronological order, if $i = k$, then $k$ is a revert, and $j$ is most likely a damaging edit. These events often occur when an article is vandalized, or when an edit does not follow the conventions of the article.[2] Not all damaging edits are immediately reverted, i.e. versions $i, j, k$ may not immediately follow each other, although they usually do.

Two categories of editors exist – anonymous and registered. Anonymous users commit most of the vandalism, although their overall legitimate contributions are rather small.[1]

For each revision, the author, editorial comment and full text of the version of the page are available.[6] This data is used to create a classifier with low false positives that can predict whether an edit by an anonymous editor is vandalism. This model can be utilized by an anti-vandal bot or other quality review software such as *Huggle*.[7]

## 3. I/O behavior

### 3.1. Input

The quality of an edit is predicted using features of the edit, its author, and the article being edited.[2] The input to the classification system will be:

- Changes to the article by the respective edit, expressed as a bag-of-words feature vector.
- Metadata associated with the edit, including:
    - The edit summary statement, also expressed as a bag-of-words feature vector.
    - The IP address of the anonymous editor.

All of the above inputs are merged into a single feature vector.

### 3.2. Output

The probabilistic classifier's output is the predicted probability of the edit being vandalism. Based on a predetermined threshold, this probability is transformed into a class, predicting whether the respective edit is vandalism or not.

## 4. Dataset

### 4.1. Raw dataset transformation

The most recent version of the required *pages-meta-history.xml.bz2* dataset[8] for the task is from March 2008. Due to technical hurdles, creation of newer versions of this dataset by Wikimedia has been halted, although newer data can be queried incrementally using a web API. The size of the compressed version of the dataset is 147 GB. A sufficiently large sample of this dataset was used. Using the entire dataset was not feasible, as the implementation of the learning algorithm requires the training data to be held in memory.

The XML dataset was parsed using a serial-access parser. Two parsers were evaluated: SAX (Simple API for XML), and *ElementTree*.[9] The latter functioned faster and requires less code.

Interlacing revisions were scanned for reverts by comparing their article texts for equality, in order to detect vandalism. As indicated earlier, because anonymous editors commit most of the vandalism, only revisions by these editors were considered for further usage. Multiple successive revisions by the same editor were merged into a single revision.

The article and comment text in each revision were tokenized into lowercased words. A custom regular expression based tokenizer was developed for this purpose. HTML tags and Wikipedia's various syntactical elements were detected as words, as these are believed to be predictive. Long words containing repetitions of smaller words, such as "hihihi…" or "funfunfun…" were detected and split into their respectively contained smaller words.

For each article, the number of instances of each unique word in each revision was determined. Using all word counts of all revisions for model development, however, would have been computationally prohibitive. Besides, only those select few words whose counts are changed from one revision to the next are believed to contain the majority of the predictive value. For these reasons, only this subset of words was considered for further use.

The differences in word counts from one revision to the next were calculated. These have values in the set of all integers. The ratio of the word counts of the two revisions was calculated as well, given that these ratios may add to the predictive value of the model. Using this ratio vector in addition to the difference vector resulted in slightly better performance, although this doubled the dataset size. These difference and ratio vectors are believed to have more predictive value than the two individual word count vectors themselves. In order to maintain and benefit from sparsity, i.e. non-representation of zero values, the ratios were transformed to be centered on zero, and be in the set of all real numbers.

While this representation results in a loss of information of the order of words, it allows text to be represented numerically, as this is necessary for many learning algorithms. Term-weighting techniques such as tf-idf (term frequency, inverse document frequency) which are commonly used were not applied, as it would result in an unacceptable loss of sparsity when computing the difference and ratio vectors. Word stemming was not deemed necessary, as different derivations of a stem may be differently predictive, given that the dataset is large.

### 4.1.1. Example

The following tables show an example demonstrating the aforementioned transformation of text into numerical vectors:

| Version | Article text |
|---------|--------------|
| $i$ | Multiverses have been hypothesized in many fields of science, including cosmology, physics, and astronomy. |
| $j$ | Multiverses have been hypothesized in cosmology, physics, astronomy, philosophy, and fiction, particularly in science fiction and fantasy. |

|              | Sparse word vectors |   |              |                    |
|--------------|---|---|--------------|--------------------|
| Word         | $i$ | $j$ | $j - i$ (Difference) | $j/i$ (Transformed ratio) |
| ,            | 3 | 5 | 2            | 1.6667             |
| .            | 1 | 1 |              |                    |
| and          | 1 | 2 | 1            | 2.0                |
| astronomy    | 1 | 1 |              |                    |
| been         | 1 | 1 |              |                    |
| cosmology    | 1 | 1 |              |                    |
| fantasy      |   | 1 | 1            | 1.0                |
| fiction      |   | 2 | 2            | 2.0                |
| fields       | 1 |   | −1           | −1.0               |
| have         | 1 | 1 |              |                    |
| hypothesized | 1 | 1 |              |                    |
| in           | 1 | 2 | 1            | 2.0                |
| including    | 1 |   | −1           | −1.0               |
| many         | 1 |   | −1           | −1.0               |
| multiverses  | 1 | 1 |              |                    |
| of           | 1 |   | −1           |                    |
| particularly |   | 1 | 1            | 1.0                |
| philosophy   |   | 1 | 1            | 1.0                |
| physics      | 1 | 1 |              |                    |
| science      | 1 | 1 |              |                    |

Positive and negative numbers in the above word vectors refer to addition and removal of words respectively. Zero values are not listed.

### 4.1.2. Ancillary features

Several potentially predictive metadata features for each revision were also computed. These are:

- a Boolean (0 or 1) indicating whether the revision has any text or is empty, as revisions with no text are likely to be vandalism
- a Boolean indicating whether the editor entered a summary comment, as one is usually not supplied for a revision that is vandalism
- a Boolean indicating whether the editor marked the revision as a minor edit, as such revisions are unlikely to be vandalism
- a Boolean indicating whether the revision was a change to the "External links" section of the article, as articles are often vandalized by adding a spam link to this section
- the first of the four octets of the IPv4 address of the anonymous editor, as different values for this correspond to different geographical regions and may be differently predictive
- the number of revisions in the collapsed sequence of successive revisions by the editor, minus one, as vandalism is likely to be just a single revision and not part of a sequence
- the number of characters and words in the previous and current revision texts and comments, and their respective differences

All of the transformed data was stored in a relational database.

## 4.2. Further transformation and scaling

Additions and subtractions of a word were processed as two separate unsigned features, instead of as one signed feature. This resulted in slightly better performance without affecting the dataset size.

Learning algorithms generally require that the values of features in the training data be in the range of 0 to 1 in order to achieve a reduced training time and optimal performance. Three scaling functions were experimented with:

| Scaling function name | Formula for $f(x)$, with $x \geq 0$ | Notes |
|---|---|---|
| atan | $\arctan(x) \times 2/\pi$ | |
| binary | $1 \text{ if } x > 0, else\ 0$ | |
| log-lin | $\ln(x+1) / \ln(\max(X)+1)$ | $\max(X)$ derived from training data. If any test $x > \max(X)$, then $f(x) = 1$. |

As required, for each function, $0 \leq f(x) \leq 1$, and $f(0) = 0$. For the *binary* function, because the difference and ratio vectors are equal after being scaled, only the difference vector needs to be used, and so the dataset size is halved.

## 4.3. Dataset summary

2,000,000 cases from the years 2001-08 were extracted from articles starting with the letters A to M. 1,585,397 unique words and an average of 104 words per case were extracted. Because additions and subtractions of words were processed as separate features, the number of features in the dataset is at least twice the number of unique words, or 3,170,794. Additionally, because *atan* and *log-lin* scaled datasets also include word ratio features, the number of features in them is at least four times the number of unique words, or 6,341,588.

43% of the cases belong to the positive class, i.e. vandalism, and the remaining 57% to the negative class, i.e. not vandalism. The baseline accuracy and RMSE on the dataset are therefore 57% and 66% respectively.

The data was split into training, validation, and test sets. 50% of the cases were randomly assigned to the training set for use by the learning algorithm. 25% were randomly assigned to the validation set, to be used for model parameter optimization. The remaining 25% were assigned to the test set, to be used only for evaluating the performance of the chosen model. Cross-validation was not used for parameter optimization due to the large size of the datasets and the prohibitive cost of training multiple models.

## 5. Model training and optimization

## 5.1. Logistic regression training

Given that the number of features in the training dataset is much larger than the number of cases, simple and highly regularized approaches are the methods of choice.[10] The Liblinear package was used to train L2-regularized logistic regression models using a trust region Newton method. Linear methods such as logistic regression tend to work well in very high-dimensional applications including document classification.[11] Regularization aids in generalization.[12] Liblinear allows for large-scale linear classification, and is efficient on large datasets.[13]

### 5.1.1. Formulation

Given data $x$, weights $(w, b)$, and class label $y$, if training instances are $x_i, i = 1, \ldots, l$ and labels are $y_i \in \{1, -1\}$, then $(w, b)$ are estimated by

$$\min_{w} f(w) \equiv \frac{1}{2} w^T w + C \sum_{i=1}^{l} \log\left(1 + e^{-y_i w^T x_i}\right)$$

where $C > 0$ is a penalty parameter.[12]

### 5.1.2. Optimization

As recommended by the authors of Liblinear, models were trained on the training set, with $C$ varied on a log scale[14] from $2^{-5}$ to $2^{11}$. A bias value $b = 1$ was used. The performance of each model was measured on both the training and the validation sets.

Since the exact misclassification costs are not known as they depend upon the application in which the model is used, the RMSE (root mean squared error) metric was optimized instead. This is because MSE tends to be correlated with profit, and so it can be used when exact costs are unavailable.[15] This allows the desired misclassification costs to be selected at a later time by varying the threshold for classification.[16]

## 5.2. Calibration

MSE can be decomposed into two components, one measuring calibration and the other measuring refinement.[17] The calibration process improves the estimate of the probability that each test example is a member of the class of interest.[18] Isotonic regression using the PAV (pair-adjacent violators) algorithm was used to calibrate each model. It is a non-parametric form of regression for reducing calibration error in the class membership probabilities returned by the logistic model.

Given predictions $f_i$ from a model and true targets $y_i$, the basic assumption in isotonic regression is that $y_i = m(f_i) + \epsilon_i$ where $m$ is an isotonic (monotonically increasing) function. Given training set $(f_i, y_i)$,

$$\hat{m} = argmin_z \sum [y_i - z(f_i)]^2$$

PAV finds the stepwise-constant isotonic function that best fits the data according to a mean-squared error criterion. It can be viewed as a binning algorithm where the position of the boundaries and the sizes of the bins are chosen according to how well the classifier ranks the examples.[15]

Even though logistic regression models tend to be reasonably calibrated by default, they are known to still benefit from calibration in most cases.[11] If the learning algorithm does not overfit the training data, the same data can be used to learn a model for calibration, without risking that this calibration model will overfit the training data.[15] Because it was observed that logistic regression did not overfit the training data greatly, specifically by an RMSE of 1.2% on average, the same training data was used to learn the calibration model.

### 5.3. Results

Experiments were performed for each of the three scaling functions. The following results were obtained:

| Scaling function | Best $C$ for RMSE | 1 – RMSE | | | | Accuracy (calibrated) | | Rank |
| --- | --- | --- | --- | --- | --- | --- | --- | --- |
| | | Training | Training (calibrated) | Validation | Validation (calibrated) | Validation | Best Threshold | |
| atan | 16 | 61.77% | 62.13% | 60.27% | **60.52%** | 78.03% | 0.5041 | 1 |
| binary | 0.125 | 61.72% | 62.1% | 60.24% | **60.49%** | 77.98% | 0.5087 | 1 |
| log-lin | 128 | 61.23% | 61.58% | 60.07% | **60.34%** | 77.72% | 0.5151 | 2 |

#### 5.3.1. 1 – RMSE

For the 1 – RMSE performance metric, higher values are better. The calibrated scores for 1 – RMSE over the validation set for the three scaling functions are comparable, with a maximum difference of less than a quarter of a percent. Calibration always improved the RMSE, although only slightly. The baseline performance for 1 – RMSE is 34.12%.

Changing $C$ had only a negligible impact on the performance on the validation set.

#### 5.3.2. Accuracy

The probability threshold for accuracy was optimized over the training set. The optimal thresholds on the calibration predictions are close to the defaults of 0.5. The learnt thresholds were then used to compute the accuracy on the validation set. Accuracy is not a useful metric by itself, since it assumes equal misclassification costs, and this assumption is rarely true. The observed accuracies are seen to be correlated with RMSE. The baseline for accuracy is 56.6%.

McNemar's Test was performed over the validation set to compare the three classifiers with each other and with the baseline. McNemar's Test is a significance test for comparing two classifiers. It is used to accept or reject the hypothesis that they have the same error rate at a given significance level.[19] Based on the test, the classifiers trained over the *atan* and *binary* scaled data have the same error rate. The classifier trained over the *log-lin* scaled data has an error rate that is different from the others.

The table below lists the computed chi-square statistics for all six pairs of classifiers, as were required for the tests. Values less than 3.84 imply the hypothesis that the two classifiers have the same error rate at significance 0.05.

| **Classifier** | Baseline | atan | binary |
| --- | --- | --- | --- |
| atan | 57,380 | | |
| binary | 57,424 | 1.01 | |
| log-lin | 56,173 | 202 | 106 |

### 5.3.3. Model selection

For 1 – RMSE, the *binary* scaling function performed almost as good as the *atan* scaling function. Because the *binary* scaling function is simpler and needs half as many features as *atan*, it results in a faster prediction time. For these reasons, *binary* was chosen over *atan*.

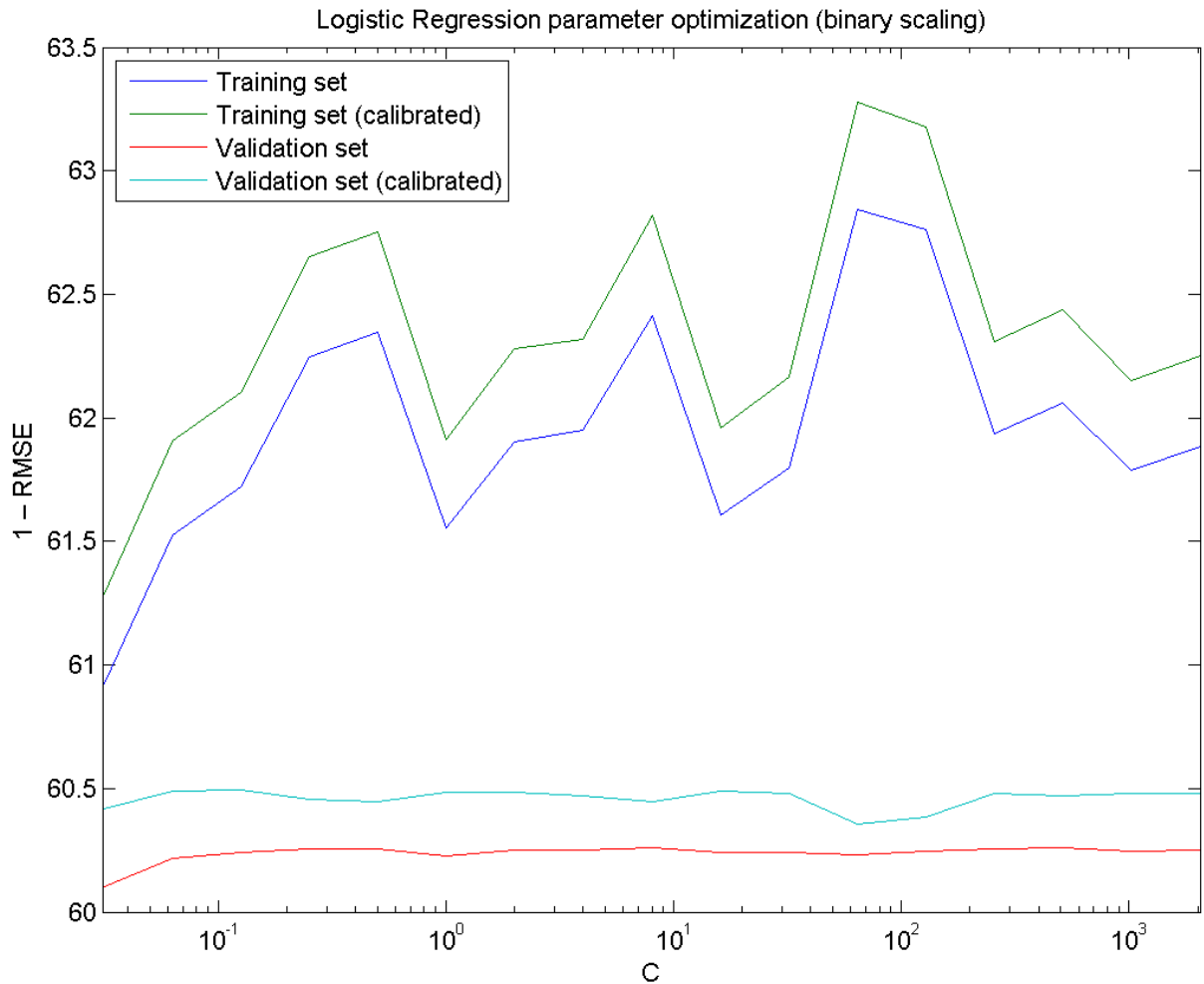

## 6. Analysis

### 6.1. Learning curve

Learning curves are used in machine learning to see the generalization performance as a function of the number of training examples. The curve is steeper for smaller training set sizes. In other words, the increase in performance is lesser for larger training set sizes.[20]

The training set size was changed on a log scale from $10^1$ to its full size of $10^6$. Performance was measured on the entire validation set. Ten iterations were performed for each training set size except the largest, in which case the full dataset was used. Training data was selected using simple random sampling without replacement.

For the learning curve, calibration using isotonic regression was not applied, as this requires at least 1,000 training cases,[15] and these were not available for all instances of the training data. The value of $C$ that previously yielded the best RMSE on the validation set without calibration (with *binary* scaling) was used.

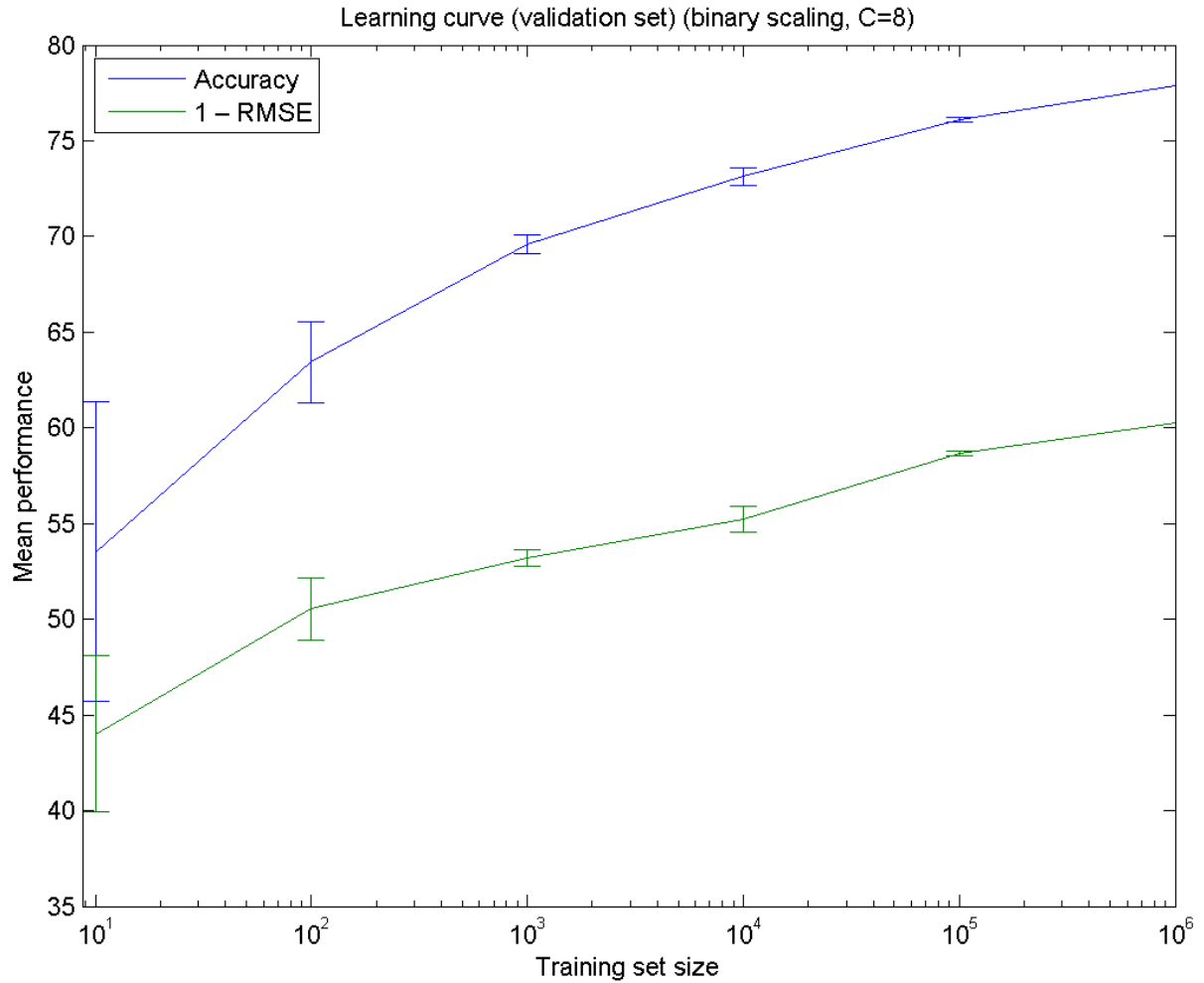

Error bars in the plot represent one standard deviation.

It is observed that the learning method does indeed learn a better model when given increasing amounts of training data. As expected, the increase in performance is lesser for larger training set sizes.

## 6.2. Reliability diagram

The calibration of a classifier can be visualized through a reliability diagram.[17] To construct a reliability diagram, the prediction space is discretized into ten bins. Cases with a predicted value between 0 and 0.1 fall in the first bin, between 0.1 and 0.2 in the second bin, etc. For each bin, the mean predicted value is plotted against the true fraction of positive cases. If the model is well calibrated the points fall near the diagonal line.[18]

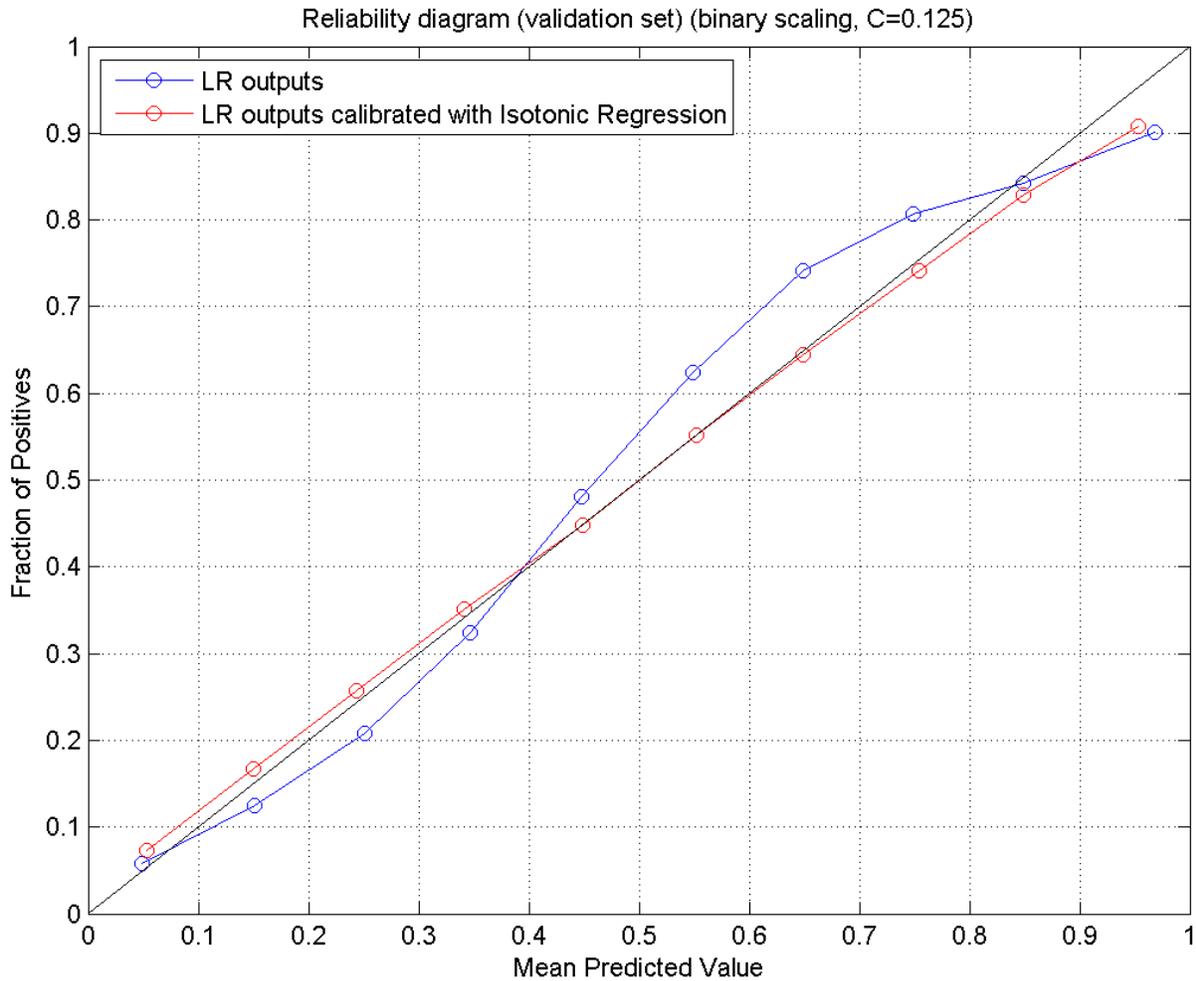

A reliability diagram was constructed over the validation set using the previously chosen model. The pre and post calibration points were plotted. As seen in the diagram, the points belonging to the calibrated classifier are closer to the diagonal.

## 7. Thresholding

Different approaches exist for selecting a threshold between 0 and 1 for classification. Predictions over the operating threshold are classified as belonging to the positive class, and the rest to the negative class.

## 7.1. ROC and PR curves

Receiver Operator Characteristic (ROC) and Precision-Recall (PR) curves show different tradeoffs achieved by varying the threshold for classification.

| Terms | Definition | Interpretation |
|---|---|---|
| TPR, sensitivity, recall | TP/(TP+FN) | Fraction of positives that are correctly classified |
| FPR, fallout | FP/(TN+FP) | Fraction of negatives misclassified as positive |
| Precision, PPV | TP/(TP+FP) | Fraction classified as positive that are truly positive |

The ROC curve shows false positive rate vs. true positive rate. PR curves show Recall vs. Precision, and are used in Information Retrieval. The goal of a learning algorithm is to be in the upper-left and upper-right corners of the ROC and PR curves respectively. The area under the curve (AUC) can be used as a metric for comparison of algorithms, with higher values being better. An algorithm that optimizes the area under the ROC curve (AUC-ROC) is not guaranteed to optimize the area under the PR curve (AUC-PR).[21]

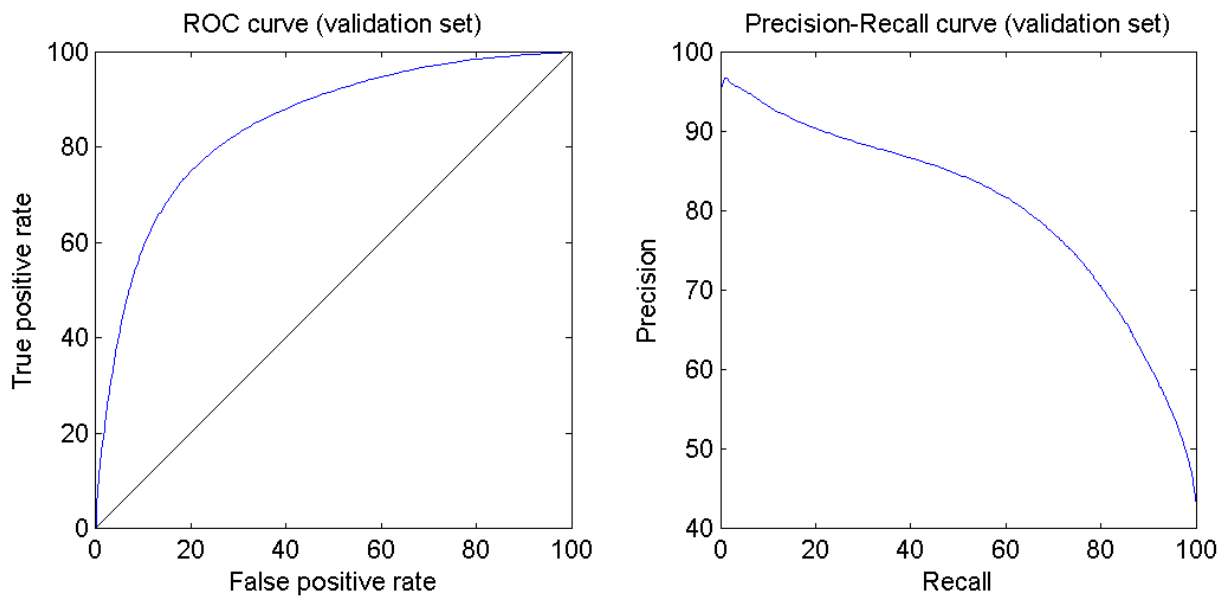

While the model was not directly optimized for AUC-ROC or AUC-PR, their values were measured for the selected model over the validation set. These were observed to be 84.73% and 80.46% respectively.

## 7.2. Cost analysis

The essence of cost-sensitive decision-making is that it can be optimal to act as if one class is true even when another class is more probable. For example, it can be rational not to approve a large credit card transaction even if the transaction is most likely legitimate.[22] As stated earlier, it is believed that a false positive would have a higher cost than a false negative in any practical application of the learning algorithm for this classification problem. The following cost matrix is proposed:

|  | actual negative | actual positive |
|---|---|---|
| **predicted negative** | $c_{00}$ = 0 (true-negative cost) | $c_{01}$ = 1 (false-negative cost) |
| **predicted positive** | $c_{10}$ >1 (false-positive cost) | $c_{11}$ = 0 (true-positive cost) |

This cost matrix meets the reasonableness condition for cost matrices, namely, that the cost of labeling an example incorrectly should always be greater than the cost of labeling it correctly.[22] Given the proposed cost matrix, effectively, only the ratio of the two misclassification costs in it is uncertain. With $c_{01} = 1$, multiple values for $c_{10}$ can be considered.

The optimal prediction is the positive class if and only if the expected cost of this prediction is less than or equal to the expected cost of predicting the negative class.[22] Given $c_{00} = c_{11} = 0$, the theoretical threshold for making an optimal decision on classifying instances as positive is $c_{10}/(c_{10} + c_{01})$. In practice however, an empirically derived threshold yields a lower cost than the theoretical threshold. The validation set can be used to search for this empirical threshold.[16]

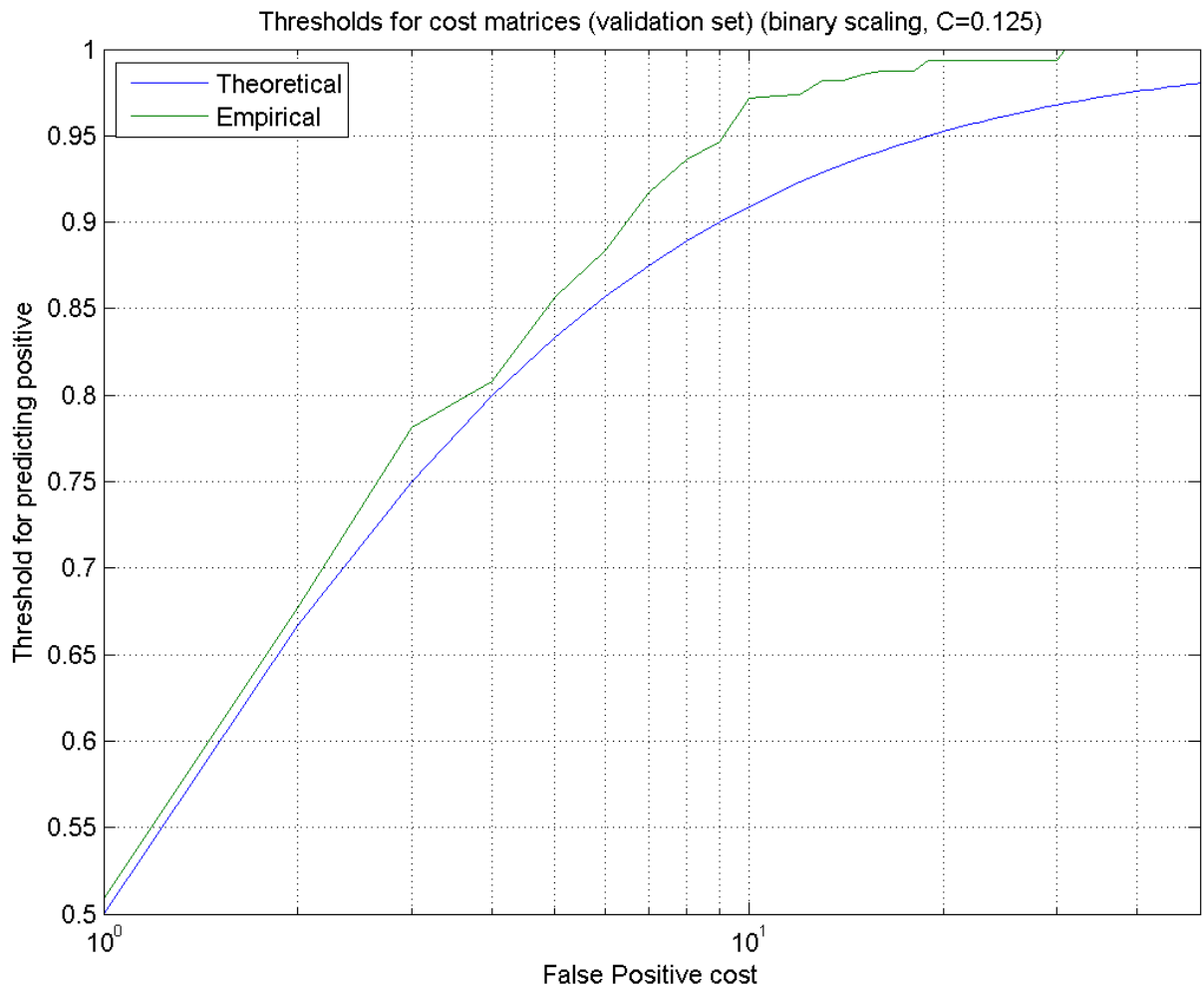

Given the suggested cost matrix with an uncertain false-positive cost, values from 1 to 50 for the false-positive cost were plotted on a log scale against the respective theoretical and empirical thresholds. It is observed that the empirical threshold is consistently higher than the theoretical. For false-positive costs greater than 30, the empirical threshold is 1, and therefore the model would not be discriminative for those costs.

## 8. Evaluation on test set

The selected model, i.e. with *binary* scaling, C = 0.125, and calibration, was evaluated on the test set. Accuracy was computed using the previously selected threshold of 0.5087. The following results were obtained:

| 1 – RMSE | Accuracy | AUC-ROC | AUC-PR |
|---|---|---|---|
| 60.47% | 77.88% ± 0.15% (99% CI) | 84.72% | 80.32% |

### 8.1. Confusion matrix

Given the previously suggested cost matrix, along with a supposed false-positive cost of 4, the empirical threshold for this specific cost matrix was previously observed to be 0.808.

A confusion matrix is a contingency table showing the differences between the true and predicted classes for a set of labeled examples.[23] The following is the confusion matrix for the aforementioned threshold on the test set, with values expressed as percents:

|  | actual negative | actual positive | sum |
|---|---|---|---|
| predicted negative | 53.84% | 25.44% | 79.28% |
| predicted positive | 2.82% | 17.91% | 20.72% |
| sum | 56.65% | 43.35% |  |

## 9. Conclusions

The L2-regularized logistic regression model with *binary* scaling, C = 0.125 and calibration using isotonic regression was the preferred model on the English Wikipedia vandalism prediction dataset.

New words are continually introduced into the Wikipedia corpus, and a failure to capture their predictive potential may lead to suboptimal predictions. The obvious solution to this concern is of course to retrain the model frequently with up-to-date data. More generally, the extent to which concept drift, i.e. a change over time in the statistical properties of the target variable, may exist in the data is unknown, and is a topic for future study.

Because there is very limited overfitting of the model to the training set, the entire dataset of 2,000,000 cases can be used to train a model using the previously estimated optimal parameters. Given sufficient memory, a further marginal improvement in performance should be possible by parsing and utilizing the entire corpus of several million cases.

Other learning methods such as linear SVM or Random Forests with calibration may perform better.[11] Using an ensemble method such as Random Forests will result in a slightly longer prediction time, however, despite opportunities for parallelization. Nonlinear SVM kernels are not believed to be feasible or useful for a dataset of this size.